\documentclass{article}
\usepackage{spconf,amsmath,graphicx}
\usepackage{multirow}

\usepackage{amsmath} 
\usepackage[subnum]{cases}

\usepackage[font=small,skip=5pt]{caption}
\title{Unsupervised Spoken Term Detection with Spoken Queries by Multi-level Acoustic Patterns with Varying Model Granularity}
\name{Cheng-Tao Chung$^{\#1}$, Chun-an Chan$^{*2}$, and Lin-shan Lee$^{\#*3}$}
\address{
Graduate Institute of Electrical Engineering, National Taiwan University$^{\#}$\\
Graduate Institute of Communication Engineering, National Taiwan University$^{*}$\\
\small
\texttt{b97901182@gmail.com$^{1}$, chunanchan@gmail.com$^{2}$, lslee@gate.sinica.edu.tw$^{3}$}
\normalsize
}
\begin{document}
%\ninept

%\fontsize{9}{9.75}\selectfont
\fontsize{8.5}{10.4}\selectfont
%%%%%%%%%%%%%%%%%%%%%%%%%
%\fontsize{9.}{30}\selectfont
%\onecolumn
%%%%%%%%%%%%%%%%%%%%%%%%%

\setlength{\textfloatsep}{2pt plus 1.0pt minus 1.0pt}
\setlength{\floatsep}{2pt plus 1.0pt minus 1.0pt}
\setlength{\intextsep}{1.2pt plus 1.0pt minus 1.0pt}
\maketitle

%\captionsetup[table]{font=small,skip=0pt}
%\captionsetup[figure]{font=small,skip=0pt}

\begin{abstract}
This paper presents a new approach for unsupervised Spoken Term Detection with spoken queries using multiple sets of acoustic patterns automatically discovered from the target corpus. The different pattern HMM configurations(number of states per model, number of distinct models, number of Gaussians per state)form a three-dimensional model granularity space. Different sets of acoustic patterns automatically discovered on different points properly distributed over this three-dimensional space are complementary to one another, thus can jointly capture the characteristics of the spoken terms. By representing the spoken content and spoken query as sequences of acoustic patterns, a series of approaches for matching the pattern index sequences while considering the signal variations are developed. In this way, not only the on-line computation load can be reduced, but the signal distributions caused by different speakers and acoustic conditions can be reasonably taken care of. The results indicate that this approach significantly outperformed the unsupervised feature-based DTW baseline by 16.16\% in mean average precision on the TIMIT corpus.%, can be comparable to that with a supervised recognizer, and combining unsupervised and supervised approaches gave by far the best performance.

%Without using any annotation, sets of acoustic patterns with different temporal and acoustic granularity were selected and discovered independently from a given corpus. The proposed method is pure data driven approach that retains most benefits of text-to-text ASR search systems such as the ability to model acoustic variation and drastically reduced search time over feature matching.
 %The results indicate that the information collected from different levels of patterns, using significantly reduced computational resources, defeats the unsupervised feature based DTW baseline and gives performance comparable to that obtained from a supervised recognizer trained with a labelled corpus. Moreover, combining unsupervised and supervised approaches gave by far the best performance.
\end{abstract}

\section{Introduction}
The fast growing quantity of video and audio content over the Internet implies a very high demand for efficient and accurate approaches to search through the spoken contents. Spoken term detection (STD) usually refers to the task of finding all occurrences of the text query term from a large spoken archive [1]. Most STD approaches were based on automatic speech recognition (ASR), transforming speech into words or subwords for token matching \cite{miller2007rapid}\cite{mamou2007vocabulary}\cite{wallace2007phonetic}\cite{pan2010performance}, with performance relying heavily on the ASR accuracy \cite{saraclar2004lattice}. This implies annotated training corpora properly matched to the spoken content are necessary. 
When the input query is spoken, it becomes possible to directly match the spoken content with the spoken query without conventional ASR. In this way, the difficulties in conventional ASR such as recognition errors and need for annotated training data may be bypassed, which is especially attractive for languages with very limited annotated data \cite{boves2009resources}\cite{kumar2007wwtw} or spoken content with unknown languages. This leads to recent efforts in unsupervised STD with spoken queries without using annotated data during training \cite{metze2012spoken}
\cite{chan2011unsupervised}, which is also the focus of this work. Hereafter we assume all queries are in spoken, and no annotated speech data is available. 

Prevailing approaches to the task considered here rely on dynamic time warping (DTW) to directly match the spoken query to the spoken documents. However, a major limitation of DTW-based approaches is that the DTW distances are easily affected by speaker mismatch and varying acoustic conditions. Many related works focused on feature representations and distance measures within the DTW framework that are more robust to speaker and acoustic condition variations\cite{carlin2011rapid}, including using the posteriorgrams from a universal Gaussian mixture model \cite{zhang2009unsupervised}, and the acoustic segment models \cite{huijbregts2011unsupervised}\cite{wang2012acoustic}. The other limitation for DTW-based approaches is that the computation load for the matching process is linear to the number of frames to be searched through. Substantial efforts were devoted to reducing this computation load, such as segment-based DTW \cite{chan2011unsupervised}, lowerbound estimation for DTW \cite{zhang2011piecewise}\cite{zhang2012fast}, and a locality sensitive hashing technique for indexing speech frames \cite{jansen2012indexing}. 

In recent years substantial effort has been made for unsupervised model based discovery of acoustic patterns from corpora without manual annotation\cite{chung2013unsupervised}\cite{jansen2010towards}\cite{jansen2011towards}\cite{gish2009unsupervised}\cite{creutz2007unsupervised}\cite{lee2013enhancing}.
% and corresponding linguistic knowledge. Most of such effort discovered only one level of phoneme-like acoustic patterns, but it is well known that there exist multi-level linguistic structures within speech signals including at least phonemes and words of different lengths. Recently we proposed to discover the hierarchical structure of two-level acoustic patterns, including HMMs for subword-like and word-like patterns, a lexicon and a language model based on which encouraging results on preliminary experiments were reported. 
In this paper, we propose a new approach for unsupervised STD with spoken queries using multi-level acoustic patterns automatically discovered from the target corpus with varying model granularity discovery from the corpus of interest. 
%It is well known that there exist various representations of speech including at least phonemes and words, thus we propose an approach for QbE STD based on unsupervised discovery of multi-level linguistic structure with HMMs of different temporal and acoustic granularity. 
The different pattern HMM configurations(number of states per model, number of distinct models, number of Gaussians per state)form a three-dimensional model granularity space. Different sets of acoustic patterns automatically discovered on different points properly distributed over this three-dimensional space are complementary to on another, thus can jointly capture the characteristics of the spoken terms. By converting the spoken content and query into parallel sequences of acoustic patterns with different model granularity, token matching can be performed with pattern indices representing highly varying signals. Very encouraging results were obtained the preliminary experiments.
\section{Acoustic Patterns with Varying Model Granularity}
\subsection{Pattern Discovery for a Given Model Configuration}

%%%
%In this section, we propose the basic concept of unsupervised discovery of parallel multi-level linguistic structure based on acoustic patterns discovered with varying temporal and acoustic granularities. 
%%%
%Given an unlabelled speech corpus, a selected length of HMM (M: number of states) and an initial set of the acoustic patterns (N: number of HMMs), it is not different to discover the desired acoustic patterns from the corpus without supervision(cite) including finding the parameter set $\theta$ of all HMMs in the transcription $W$ for the observed acoustic feature vector $\bar{O}$ of the corpus based on the HMMs described by $\theta$. 
Given an unlabelled speech corpus, it is not difficult for unsupervised discovery of the desired acoustic patterns from the corpus for a chosen hyperparameter set $\psi$ that determines the topology of the HMMs \cite{jansen2011towards}\cite{gish2009unsupervised}\cite{creutz2007unsupervised}.
This can be achieved by first finding an initial label $\omega_0$ based on a set of assumed patterns for all observations in the corpus $\chi$ as in (\ref{eq:1})\cite{chung2013unsupervised}.
%which can be done with some clustering approach such as the hierarchical agglomerative clustering (HAC) algorithm(cite).
Then in each iteration $t$ the HMM parameter set $\theta^\psi_{t}$ can be trained with the label $\omega_{t-1}$ obtained in the previous iteration as in (\ref{eq:2}), and the new label $\omega_{t}$ can be obtained by free-pattern decoding with the obtained parameter set $\theta^\psi_{t}$ as in (\ref{eq:3}). 
\begin{eqnarray}
\omega_{0}&=& \mbox{initialization}(\chi),                                           \label{eq:1} \\ 
\theta^\psi_{t} &=& \arg \max_{\substack{\theta^\psi}} P(\chi|\theta^\psi,\omega_{t-1}),             \label{eq:2} \\
\omega_{t} &=& \arg \max_{\substack{\omega}} P(\chi|\theta^\psi_{t} ,\omega).                        \label{eq:3}
\end{eqnarray}
The training process can be repeated with enough number of iterations until the difference between $\omega_{t-1}$ and $\omega_{t}$ becomes insignificant. This gives a converged set of acoustic pattern HMMs which we denote as $\Theta^\psi$. 
\subsection{Model Granularity Space}
\begin{figure}[tbh]
%\centerline{\epsfig{figure=figure,width=80mm}}
\centerline{\includegraphics[width=0.4\textwidth]{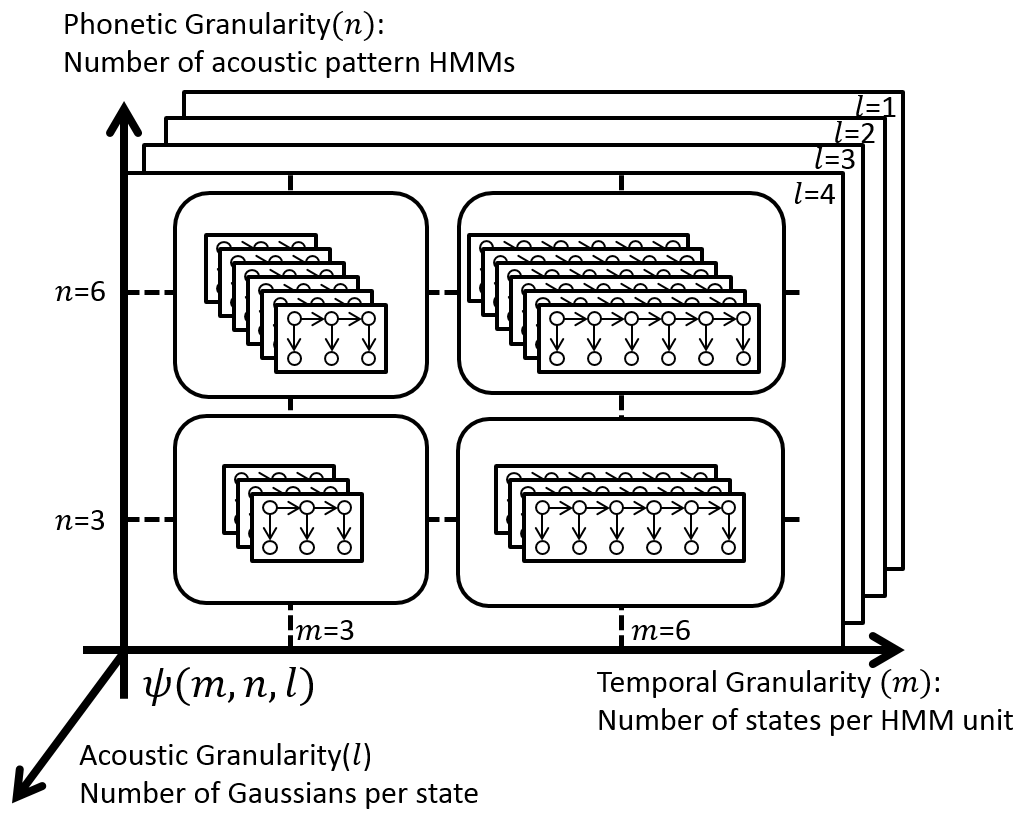}}
\caption{The model granularity space for acoustic pattern configurations.}\label{fig:cube}
\end{figure}

The above process can be performed with many different HMM configurations, each characterized by three hyperparameters: the number of states $m$ in each acoustic pattern HMM, the total number of acoustic patterns $n$ during initialization, and the number of Gaussians $l$ in each HMM state, $\psi=(m,n,l)$. The transcription of a speech signal decoded with these acoustic pattern HMMs may be considered as a temporal segmentation of the signal, so the HMM length (or number of states in each HMM) $m$ represents the temporal granularity. The set of all acoustic pattern HMMs may be considered as a segmentation of the phonetic space, so the total number $n$ of acoustic pattern HMMs represents the phonetic granularity. The different Gaussians in each state then jointly model the distributions of the signals in the acoustic feature space represented by MFCCs, so the number of Gaussians $l$ in each state represents the acoustic granularity.
This gives a three-dimensional representation of the acoustic pattern configurations in terms of temporal, phonetic and acoustic granularities as in Fig. \ref{fig:cube}. Any point in the three-dimensional space in Fig. \ref{fig:cube} corresponds to an acoustic pattern configuration.
%This gives a three-dimensional representation of the acoustic pattern configurations in terms of temporal, phonetic and acoustic granularities as in Fig. \ref{fig:quant}, in which $m$=3 and 13, and $n$=50 and 300, and $l$=1 are shown as $\psi_k$, $k=1,2,3,4$. Any point on the three-dimensional space in Fig. \ref{fig:quant} corresponds to an acoustic pattern configuration.
%

%First look at the $(m,n)$ plane for $l=1$ as shown in Fig \ref{fig:quant}.
%Pattern sets $\psi_k$ in different areas of this plane may exhibit the following behaviour:
%\begin{itemize}
%\itemsep 0mm
%\item top-left(e.g.$\psi_2$):     finer temporal and phonetic granularities, may lead to over-fitting with insufficient training data
%\item bottom-right(e.g.$\psi_3$): coarser temporal and phonetic granularities, may lead to under-fitting with each temporal or phonetic segmentation covering too much variation
%\item top-right(e.g.$\psi_4$):    longer temporal segments with finer phonetic segments, possibly higher level patterns closer to semantic units
%\item bottom-left(e.g.$\psi_1$):  shorter temporal segments with coarser phonetic segments, possibly lower level patterns closer to signal fragments with wider variation 
%\end{itemize}
%The third dimension $l$ for acoustic granularity in Fig \ref{fig:quant}. then gives different number of Gaussians in each HMM state. Increased value of $l$ implies each HMM state may cover a wider range of signal variations in both temporal and phonetic dimensions, or more flexibilities in the segmentation of these two dimensions.

Although the hyperparameters $\psi=(m,n,l)$ are difficult to determine for a given corpus of unknown language and unknown linguistic characteristics, it is possible to have many different sets of acoustic patterns with hyperparameters and HMMs \{$\Theta^{\psi_k}, \psi_k=(m_k,n_k,l_k),k=1,2,...K$\} independently learned in parallel, considered as on different levels. Because different model granularities $(m,n,l)$ give different characteristics to the acoustic patterns as mentioned above, with enough number of pattern sets and the model granularities $(m,n,l)$ properly distributed in the three-dimensional space, this multi-level set of acoustic pattern may jointly represent the behavior of the signal for the given corpus. There can be a variety of applications for these patterns, but here in the work below we use these patterns in spoken term detection, assuming the characteristics of the spoken terms can be captured by different sets of these patterns.

%In the previous work \cite{first}, we proposed an unsupervised approach to discover the hierarchical two-level linguistic structure, which included the subword-like and word-like acoustic pattern HMMs, a lexicon, a language model. The many parameters of that structure are heavily dependent on each other in a hierarchical way, so the unsupervised discovery of such hierarchical linguistic structure may not be robust enough across different corpora of different languages with different quality. In contrast, in this approach proposed here, different levels of acoustic patterns $\psi_k=(m_k,n_k,l_k)$  with different model granularities are independently discovered in parallel, so they are complementary to and less dependent of each other. They are more reliable and easier to train, and the number of levels can be significantly more than two. 

\section{Spoken Term Detection and search methods}
%The task of spoken term detection (STD) here is to return a list of spoken documents in th archive which contains the spoken query. 
\subsection{Off-line Processing}
All the spoken documents in the archive are first off-line decoded into sequences of acoustic patterns using each level of acoustic pattern HMM set $\Theta^{(m_k,n_k,l_k)}$. Let \{$p_r, r=1,2,3,..,n_k$\} denote the $n_k$ acoustic patterns in the set $\Theta^{\psi_k}$. We further construct a similarity matrix $S$ of size $n_k \times n_k$ for every $\psi_k$, for which the component $S(i,j)$ is the similarity between any two pattern HMMs $p_i$ and $p_j$ in the set $\Theta^{\psi_k}$. Two similarity matrices used for this work are in (\ref{eq:4}).
%\begin{equation}
%S^k_{eq}(i, j) = \delta(i,j),
%\end{equation}
%\begin{equation}
%S^k_{kl}(i,j) = \mbox{exp}(-\mbox{KL}(i, j)/\beta).
%\end{equation}
%\begin{equation}
%S(i, j) = \begin{cases} \delta(i,j), & \mbox{for }             \mbox{no KL} \\ 
%\mbox{exp}(-\mbox{KL}(i, j)/\beta, & \mbox{for }         \label{eq:4} \mbox{KL divergence} \end{cases} 
%\end{equation}
\begin{subnumcases}
{S(i, j) = \label{eq:4}} \delta(i,j), & \mbox{for }             \mbox{hard similarity, or }   \label{eq:4a}\\ 
\mbox{exp}(-\mbox{KL}(i, j)/\beta, & \mbox{for }     \mbox{soft similarity.}  \label{eq:4b}
\end{subnumcases}
 The matrix in (\ref{eq:4a}) is simply the identity matrix. The KL-divergence $\mbox{KL}(i,j)$ between two pattern HMMs in (\ref{eq:4b}) is defined as the KL-divergence between the states based on the variational approximation \cite{hershey2007approximating} summed over the states. To transform the KL divergence into a similarity measure between 0 and 1, a negative exponential was applied \cite{marszalek2008constructing} with a scaling factor $\beta$. When $\beta$ is small, similarity between distinct patterns in (\ref{eq:4b}) approaches zero, so (\ref{eq:4b}) becomes similar to (\ref{eq:4a}). $\beta$ can be determined with a held out data set, but here we simply set it to 100 times the number of states $m_k$. 

\subsection{On-line Matching Matrix Construction}

\begin{figure}[tbh]
%\centerline{\epsfig{figure=figure,width=80mm}}
\centerline{\includegraphics[width=0.3\textwidth]{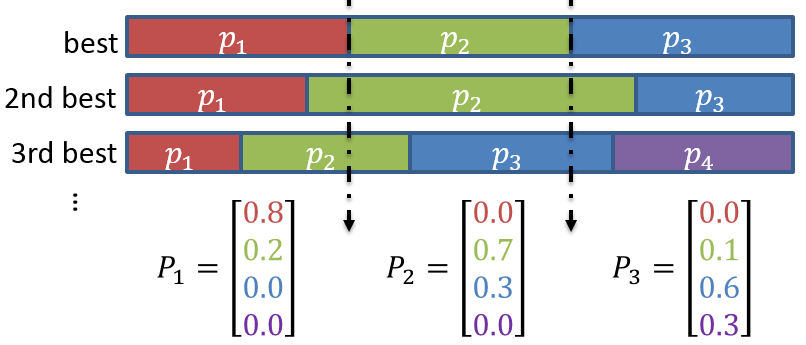}}
\caption{Construction of pattern posteriorgram from N-best list}\label{fig:lattice}
\end{figure}

In the on-line phase, we perform the following for each entered spoken query $q$ and each document $d$ in the archive. Assume for the pattern set $\Theta^{\psi_k}$ a document $d$ is decoded into a sequence of $D$ acoustic patterns with indices $(d_1, d_2, ..., d_D)$ and the query $q$ into a sequence of $Q$ patterns with indices $(q_1, ..., q_Q)$. 
We thus construct a matching matrix $W$ of size $D \times Q$ for every document-query pair, in which each entry $(i,j)$ is the similarity between acoustic patterns with indices $d_i$ and $q_j$ as in (\ref{eq:5a}).
% and shown in Fig \ref{fig:warp}(a) for a simple example of $Q=3$ and $D=6$, where $S$ can be those either in (\ref{eq:4a}) or (\ref{eq:4b}). 
%\begin{equation}
%W(i, j) = S^k(d_i, q_j),
%\end{equation} 
%\begin{equation}
%W(i, j) = \begin{cases} S(d_i, q_j), & \mbox{if }         \mbox{no %posteriorgram} \\
%P_{d_i}^T\times S\times P_{q_j}, & \mbox{if }        \label{eq:5} %\mbox{posteriorgram} \end{cases}
%\end{equation}
\begin{subnumcases}
{W(i, j) = \label{eq:5}} S(d_i, q_j), & \mbox{for }             \mbox{1-best sequences, or}   \label{eq:5a}\\ 
P_{i}^T S P_{j}, & \mbox{for }     \mbox{N-best sequences.}  \label{eq:5b}
\end{subnumcases}
%Instead of considering only the best sequence from both $q$ and $d$, 
Alternatively, we can also match the N-best sequences of documents to N-best sequences of queries as depicted in Fig (\ref{fig:lattice}) and (\ref{eq:5b}).  We extend each pattern position $i$ in a sequence into a posteriorgram vector $P_i$ of size $n_k\times 1$ by accumulating the duration for each of the $n_k$ patterns within the pattern boundaries of the best transcription, across the N-best transcriptions considered, uniformly weighted and normalized($P_i$ for $d$ and $P_j$ for $q$). When only the best transcription is considered, (\ref{eq:5b}) reduces to (\ref{eq:5a}).  The price paid here is the increased computation time by a factor of $O(n_k^2)$ over that in (\ref{eq:5a}). One can also choose to use the posteriorgram only on the query, which increases the load by $O(n_k)$.
% or choose to only compute the non-zero entries in $P_i$ and $P_j$ which only increases the load by at most $O(N^2)$ where $N$ is the number of transcriptions con.
%\begin{equation}
%R(d,q) = \sum_{k=1}^{K} \sum_{t=1}^{T} \lambda(k;t)  R_{t}^k(d,q) .
%\end{equation}

%\begin{equation}
%\lambda(k;t) = \lambda(m,n,l;a,b,c)
%\end{equation}

%This method does not consider the inevitable insertion, deletion or substitution during decoding, therefore refered to as simple matching here. 

\subsection{On-line Matching Policy}
\begin{figure}[tbh]
%\centerline{\epsfig{figure=figure,width=80mm}}
\centerline{\includegraphics[width=0.4\textwidth]{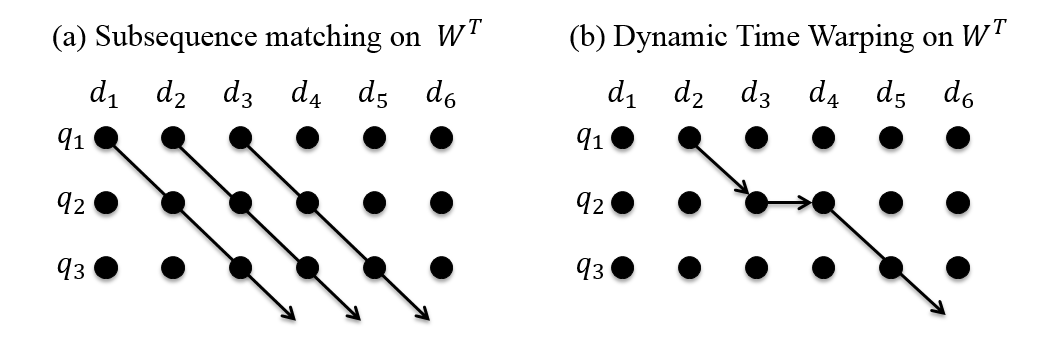}}
\caption{The matching matrix $W$}\label{fig:warp}
\end{figure}

There can be two methods to calculate the relevance score between document $d$ and query $q$. In the sub-sequence matching(SUB) method in (\ref{eq:6a}) , we sum the elements in the matrix $W$ along the diagonal direction, generating the accumulated similarities for all sub-sequences starting at all pattern positions in $d$ as shown in Fig (\ref{fig:warp})(a). The maximum is selected to represent the relevance between document $d$ and query $q$ on the pattern set $\Theta^{\psi_k}$ as in (\ref{eq:6a}).
\begin{subnumcases}
{R(d,q) = \label{eq:6}} 
\max_{\substack{i:\mbox{index} }}\sum_{j=1}^{Q} W(i+j,j)
, & \mbox{for }   \mbox{SUB, or}   \label{eq:6a}\\ 
\max_{\substack{u:\mbox{ path} }} \sum_{j=1}^{|u|} W(u_d(j),u_q(j) )
, & \mbox{for }     \mbox{DTW.}  \label{eq:6b}
\end{subnumcases}
%\begin{equation}
%R_{sm}^k(d,q) = \max_{\substack{i =0,...,D-Q}}\sum_{j=1}^{Q} W(i+j,j).
%\end{equation}
In order to alleviate the problem of insertion/deletion, we can also perform dynamic time warping (DTW) on the matrix $W$ as in (\ref{eq:6b}) and Fig \ref{fig:warp}(b), refered to as the pattern-based DTW here. 
This is an extended version of (\ref{eq:6a}), 
%\begin{equation}
%R_{dtw}^k(d,q) = \max_{\substack{u:\mbox{DTW path} }} \sum_{}^{} W(i,j).
%\end{equation}
except now the elements $W$ are accumulated along any allowed DTW path in the matrix $W$. The summation in (\ref{eq:6b}) is over a single DTW path, while the maximization in (\ref{eq:6b}) is over all DTW paths. Although DTW takes longer time, performing DTW in the matrix $W$ on-line is significantly faster than the conventional frame-based DTW, because most of the calculation was performed offline when evaluating (\ref{eq:4}). Since table lookup is constant time operation, asymptotically speaking, the computational load online is reduced by a factor of $O(FT^2)=O(Fm_k^2)$, where $F$ is the feature dimension in the frame-based DTW and $T$ is the duration of the acoustic patterns in frames which scales linearly with number of states $m_k$ in HMMs. 

\subsection{Overall Relevance Score}
In each of (\ref{eq:4})(\ref{eq:5})(\ref{eq:6}) there are two options, leading to a total 8 search methods. We thus use three binary digits to specify these methods in reporting experimental results below: $\gamma=(\mbox{Soft,Nbest,DTW})$, i.e. Soft=1 for soft similarity in (\ref{eq:4b}) and 0 for (\ref{eq:4a}); Nbest=1 for N-best sequence in (\ref{eq:5b}) and 0 for (\ref{eq:5a}); DTW=1 for DTW in (\ref{eq:6b}) and 0 for (\ref{eq:6a}). These search methods,\{$\gamma_s,s=1,...,8$\} give different relevance scores for each pattern set $\Theta^{\psi_k}$, $R^{(\psi_k,\gamma_s)}(d,q)$ as in (\ref{eq:6}).
%This would result in different similarity scores for the same pattern set $\Theta^{\psi_k}$. Therefore, the relevance score $R$ depends on both of HMM pattern parameter $\psi$ and search method $\gamma$.
The overall relevance score $\bar{R}(d,q)$ between $d$ and $q$ is then simply the weighted sum of (\ref{eq:6}) over all $K$ different sets of acoustic patterns $\Theta^{\psi_k}$ and the 8 search methods with weights $\lambda({\psi_k},{\gamma_s})$ as in (\ref{eq:sum}). Below, we simply set the weights $\lambda(\psi_k,\gamma_s)$ to either 0 or 1. 
\begin{equation}
\bar{R}(d,q) = \sum_{k=1}^{K} \sum_{s=1}^{8} \lambda(\psi_k,\gamma_s)  R^{(\psi_k,\gamma_s)}(d,q). \label{eq:sum}
\end{equation}
Note that the acoustic pattern sets $\{\Theta^{\psi_k}, k=1,2,...,K\}$ for different model granularities are complementary to one another. By adding the scores obtained via different pattern sets as in (\ref{eq:sum}) the signal characteristics can be better captured. In addition, the limitation caused by temporal granularity (model length $m_k$) may be alleviated to some degree by the pattern-based DTW in (\ref{eq:6b}), the limitation caused by phonetic granularity (number of patterns $n_k$) may be alleviated to some degree by the N-best sequences in (\ref{eq:5b}), and the limitation caused by acoustic granularity (number of Gaussian $l_k$) may be taken care to some degree by the soft similarity in (\ref{eq:4b}). Therefore, the choices of $\psi_k$ and $\gamma_s$ are correlated for good scores of $\bar{R}(d,q)$ in (\ref{eq:sum}). Also, the process above, including constructing the KL-divergence matrices and decoding the spoken documents into acoustic patterns off-line, and generating the pattern sequence for the query on-line can all be performed in a highly parallel manner, so the computational load can be scalable with the computational resources available. 
%as the performance metric. Given a score metric, for every query , we sort the document relevance score in descending order and compare it to the true relevant answer list. Starting at the start of the list, the precision at every true answer is averaged, giving us the average precision. Once again averaging the average precision across all queries gives us the mean average precision\cite{}. 

\section{Experiments}
The proposed approach was tested in preliminary experiments performed on the TIMIT corpus. 20 sets of acoustic patterns with number of states $m$=${3,5,7,9,11}$ and number of HMMs $n$=${50,100,200,300}$ were first generated with $l$=$1$ Gaussian per state on the TIMIT training set. Then we increased the number of Gaussians per state by 1 and perform (\ref{eq:2}) and (\ref{eq:3}) until the sets converge. We repeated the process until we had $l$=${1,2,3,4}$ for all the 20 pattern sets, obtaining $K$=$80$ pattern HMM sets in the end. With the 8 search methods mentioned in section 3, this gave a total of 640 scores for every query-document pair in (\ref{eq:sum}). The TIMIT training set was also taken as the spoken archive from which we wish to detect the spoken terms. 

The query set consisted of 32 spoken words randomly selected from the TIMIT testing set. An instance of every query word was randomly selected from the testing set, and used as the spoken query to search for other instances in the training set. 
%\cite{zhang2005investigations}
Note that although choices of $\lambda(\psi_k,\gamma_s)$ can be based on the optimization with respect to an evaluation metric \cite{radlinski2005query}\cite{liu2009learning}, the preliminary experiments in this work were to verify the feasibility of the proposed frameworks instead. 
The baseline we compared to was frame-based DTW on MFCC sequences. The same acoustic features were used for training the pattern HMMs. In principle, the framework should generalize to other features as well. 
%Due to recent breakthroughs, deep neural networks have been integrated into many traditional systems. Therefore, we plan to replace MFCCs with bottleneck features extracted from stacked denoising auto-encoders \cite{vincent2010stacked} in the future.
For spoken term detection the performance measures we used were the mean average precision(MAP), precision at 5 and 10(P@5 and P@10). All three measures gave very similar trends. Below we only report results for MAP due to space limitation.

% We are well aware of the abundance of ensemble methods \cite{shinozaki2010unsupervised} or probabilistic graphical approaches \cite{gael2008infinite}\cite{glass} that are suitable for the desired tasks. However, rather than pursuing performance by optimizing with respect to an evaluation metric, the experiments explored in this work focuses more on proving the feasibility of the proposed frameworks. Although models in this work were trained on MFCC features, in principle, the framework should generalize to other features as well. Due to recent breakthroughs, deep neural networks have been integrated into many traditional systems. Therefore, we plan to replace MFCCs with bottleneck features extracted from stacked denoising auto-encoders \cite{vincent2010stacked} in the future.

\subsection{Feature Selection and Achievable Performance}
\begin{figure}[tbh]
%\centerline{\epsfig{figure=figure,width=80mm}}
\centerline{\includegraphics[width=0.45\textwidth]{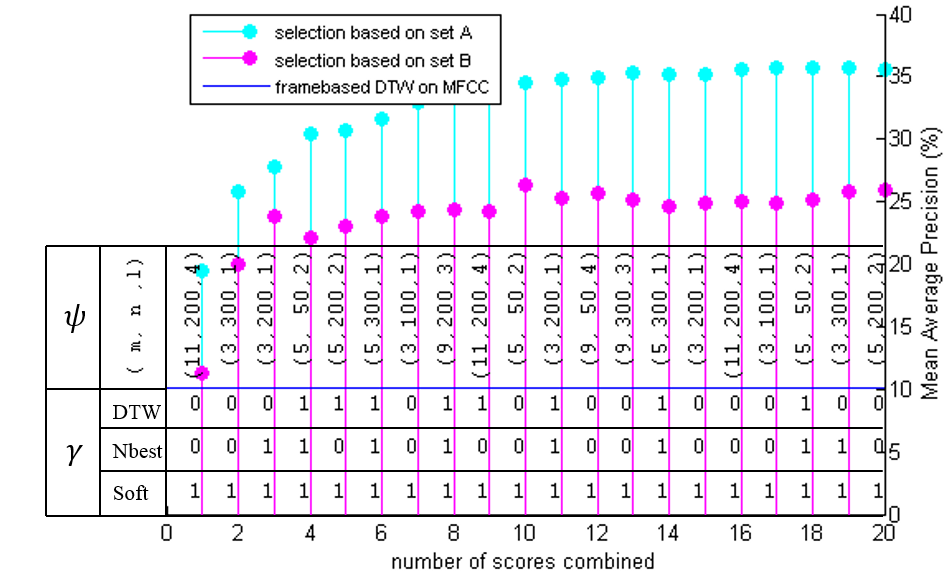}}
\caption{MAP performance on query set A with top 20 scores selected from a developement set.}\label{fig:combine}
\end{figure}

In this experiment, our goal was to learn the 640 weights $\lambda(\psi,\gamma)$ in (\ref{eq:sum}) to be either 1 or 0 in order to optimize the MAP. We randomly split the query set into 2 disjoint sets A and B, each containing 16 queries. We use set B as the development set for learning these weights and focus our discussion on the performance on set A. Starting with every $\lambda(\psi,\gamma)$=0, we greedily select the next $(\psi,\gamma)$ that would yield the best MAP and set it to 1. This process was repeated until 20 pairs of $(\psi,\gamma)$ were selected. The results are shown in pink in Fig. \ref{fig:combine}, in which the oracle results by learning on set A itself are also shown in cyan. The baseline of frame-based DTW was 10.16\% for set A. 
It was low probably because most queries unexpectedly had less than 10 relevant documents in the training set, and the various dialects of TIMIT made it even more difficult for simply comparing the feature sequences. 
We can clearly see with only 20 out of 640 scores selected based on set B, the MAP reaches 25.88\%, significantly higher than the baseline. The oracle results learning on set A itself reached as high as 35.60\%, which implied that if a more elaborate learning algorithm was applied, there was still much room for improvement. 

The parameter sets $(\psi,\gamma)$ for the 20 scores selected based on set B are also printed on Fig. \ref{fig:combine}. Some observations can be made here. In all the 20 cases Soft=1, so Soft=1 was certainly better. 
%(soft similarity using KL-divergence)
For the selection of N-best sequences or DTW there were no clear trends. However, some correlation between search methods and pattern configurations can be observed. For example, it can be found that Nbest=1 were preferred when $l$=$1$ or $l$=$2$, probably because too few number of Gaussians limited the accuracies in the 1-best sequence. Also, DTW=1 was preferred very often when $m$=$5$ or $3$, probably because for shorter patterns it made better sense to merge more than one patterns into a longer pattern, which is actually what the pattern-based DTW did.
These results seem to imply the need to learn from a development set, which is not always available. 
This will be  further discussed in the next section.

\subsection{Performance Analysis without a Development Set}
\begin{figure}[tbh]
%\centerline{\epsfig{figure=figure,width=80mm}}
\centerline{\includegraphics[width=0.5\textwidth]{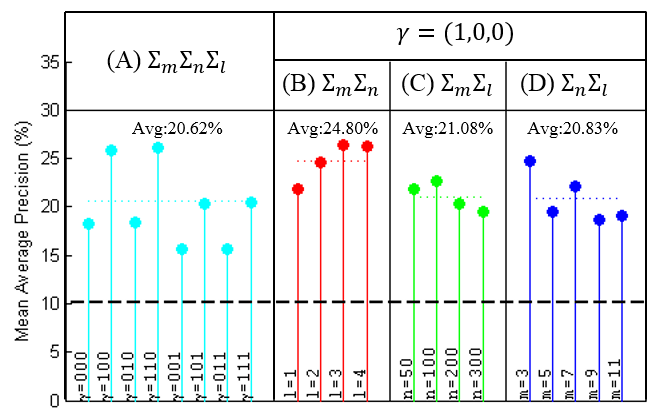}}
\caption{MAP performance on query set A without a developement set}\label{fig:dim}
\end{figure}

Here we consider the case without a development set.
We first consider the different search method $\gamma$ by summing all the 80 scores for all combinations of $m$,$n$,$l$, but not $\gamma$ as shown in Part (A) of Fig. (\ref{fig:dim}). Several conclusions may be drawn: (a) Soft similarity brought massive improvement (Soft=1 $>$ Soft=0 for all combinations of Nbest and DTW), (b) N-best Sequences brought only a negligible improvement (Nbest=1 $\sim$ Nbest=0 for all combinations of Soft and DTW), (c) DTW degraded the performance in general (DTW=1 $<$ DTW=0 for all combinations of Soft and Nbest). 
Conclusion (a) is consistent with the conclusion drawn from the top 20 scores in Fig \ref{fig:combine}. Since generating a soft similarity metric is also the only part of the search that could be conducted off-line, it is certainly attractive.
Conclusion (b) and (c) may be a little surprising, since intuitively N-best sequences and DTW can help and quite several of the top 20 scores in Fig \ref{fig:combine} had Nbest=1 or DTW=1. As discussed previously with Fig \ref{fig:combine}, Nbest and DTW could be helpful for specific pattern configurations $\psi$ but not necessarily all. When such specific configurations cannot be properly chosen with a development set, these improvements could be diluted when averaged with all possible configurations. Because the different pattern sets carried complementary information, jointly considering the 80 1-best sequences obtained from the 80 pattern sets itself can be viewed as considering a single large lattice best representing the utterance with all time warping and N-best information included. This is probably why N-best and DTW didn't help here. The MAP obtained by summing all 640 scores without any selection was 20.62\% as shown in Part(A) of Fig (\ref{fig:dim}), which was also significantly higher than the baseline. 

Therefore below we focus our discussion on the selection of pattern configurations of $\psi$ assuming $\gamma$=(1,0,0) without using Nbest or DTW. The 80 scores for different $\psi$ forms a 3 dimensional space over $(m,n,l)$. We summed the relevance score over 2 of the dimensions and plotted the performance on the remaining dimension. Summing over $(m,n)$, $(m,l)$, $(n,l)$, we get Parts (B)(C)(D) of Fig. (\ref{fig:dim}) respectively. The average of the MAP values for Parts (B)(C)(D) of Fig. (\ref{fig:dim}) are also listed for comparison between the dimensions. 
Several additional conclusions may be drawn: (d) the performance is better for larger $l$ as shown in Part(B) of Fig. \ref{fig:dim}, (e) Model combination on the $(m,n)$ plane was the most effective (the average MAP of Part(B) was much higher than those in Part (C)(D) of Fig. \ref{fig:dim}), (f) several optimal values seem to exist for $m$ and $n$ (the maximum occurs for $m$=3 in Part (D) and $n$=100 in Part (C)).
We highly suspect that these optimal points for $m$, $n$ as mentioned in Conclusion (f) are inert characteristics of the underlying language that should stay approximately the same for a different corpus of the same language. 
%
%%%%%%%%%%%%%%%%%%%%%%%%%%%%%%%%%%%%%%%%%%%%%%
Conclusion (d) will probably hold until the phenomenon of over-fitting begins to happen, since the number of Gaussians is limited by the training corpus size. 
Conclusion (e) implies when blending score with respect to $(m,n)$, a good policy may be to have $m$ and $n$ as diverse as possible, if there is no information regarding the selection of $m$, and $n$. 
From conclusion (f), $m$=3 seemed close to the average phoneme duration of TIMIT, while $n$=100 seemed close to the number of phonemes with some context dependency considered. This may imply selection of $(m,n)$ is a language dependent characteristic although we do not have strong evidence yet to back up this claim. 
This would be useful if verified to be true, especially for under-resourced languages, since in that case the learned weights could be similarly useful for different corpora on the same language.
%
%obvious if we take the average of the MAPs on the 3 dimensions respectively. The average performance by blending scores with respect to $(m,n)$ is yields a higher value than $(m,l)$ or $(n,l)$. When selecting $\psi_k$ it is generally a good idea to have $m_k$ and $n_k$ as diverse as possible. Conclusion (f) may seem like a leap of faith as we do not have figures to back up this claim, but if it is indeed true, then it would imply that the learned weights $\lambda$ can be applied to other corpora of the same language. This would be incredibly useful, especially for under resourced languages, since the learning only has to be done once for every language. 
%

\begin{figure}[tbh]
%\centerline{\epsfig{figure=figure,width=80mm}}
\centerline{\includegraphics[width=0.5\textwidth]{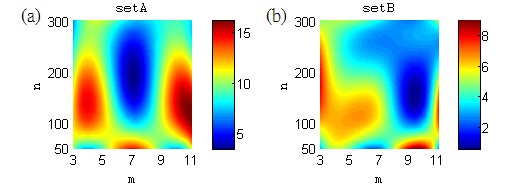}}
\caption{MAP for $\gamma$=(1,0,0) and $l$=3 over the $(m,n)$ plane for query sets A and B. }\label{fig:spline}
\end{figure}

We further plot the MAP of $R^{(\psi,\gamma)}(d,q)$ for $\gamma$=(1,0,0) and $l$=3 for sets A and B in Fig \ref{fig:spline}, the best performing $\gamma$ and $l$ in Parts (A)(B) of Fig \ref{fig:dim}. The 20 points for $(m,n)$ were interpolated with a 2D spline function to show the smoothed MAP distributions over the plane. As can be seen, the score distributions look similar to a good degree even for completely different query sets in Fig \ref{fig:spline}(a) and (b). When we simply selected the 20 scores for set A here ($m$=3,5,7,9,11; $n$=50,100,200,300; $l$=3,$\gamma$=(1,0,0)), the MAP is 26.32\%, slightly higher than 25.88\% achieved above by the 20 scores selected greedily from a development set. 

%In order to support Conclusion(f), we plot the performance of the $(m,n)$ plane for the best $l$ for query sets A and B as shown in Fig \ref{fig:spline}. There were only 20 data points originally, but we interpolated the surface with the 2D spline function to better show the invariance of the optimal regions with respect to query set.

\section{Conclusion} 
In this work, we present a new approach for unsupervised spoken term detection using multi-level acoustic patterns discovered from the target corpus. The different pattern sets with different model configurations are complementary, thus can jointly capture the information for the spoken terms. Significantly better performance than frame-based DTW on TIMIT corpus as obtained.

\bibliography{strings,refs}
\include{refs}
\bibliographystyle{IEEEbib}

\ninept

\end{document}